# PolSAR Image Classification Based on Dilated Convolution and Pixel-Refining Parallel Mapping Network in the Complex Domain

Dongling Xiao, Chang Liu, Qi Wang, Chao Wang, Xin Zhang

*Abstract*—Efficient and accurate polarimetric synthetic aperture radar (PolSAR) image classification with a limited number of prior labels is always full of challenges. For general supervised deep learning classification algorithms, the pixel-by-pixel algorithm achieves precise yet inefficient classification with a small number of labeled pixels, whereas the pixel mapping algorithm achieves efficient yet edge-rough classification with more prior labels required. To take efficiency, accuracy and prior labels into account, we propose a novel pixel-refining parallel mapping network in the complex domain named CRPM-Net and the corresponding training algorithm for PolSAR image classification. CRPM-Net consists of two parallel sub-networks: a) A transfer dilated convolution mapping network in the complex domain (C-Dilated CNN) activated by a complex cross-convolution neural network (Cs-CNN), which is aiming at precise localization, high efficiency and the full use of phase information; b) A complex domain encoder-decoder network connected parallelly with C-Dilated CNN, which is to extract more contextual semantic features. Finally, we design a two-step algorithm to train the Cs-CNN and CRPM-Net with a small number of labeled pixels for higher accuracy by refining misclassified labeled pixels. We verify the proposed method on AIRSAR and E-SAR datasets. The experimental results demonstrate that CRPM-Net achieves the best classification results and substantially outperforms some latest state-of-the-art approaches in both efficiency and accuracy for PolSAR image classification. The source code and trained models for CRPM-Net is available at: *https://github.com/PROoshio/CRPM-Net*.

## I. Introduction

SYNTHETIC Aperture Radar (SAR) is widely used in terrain classification, target detection and image change detection thanks to its all-weather, all-time and high-resolution imaging capabilities. Compared with single-polarized SAR, PolSAR, a multi-channel SAR system, has more abundant polarimetric target decomposition features and phase information. Hence it has received extensive attention in the study of terrain classification.

In general, the traditional supervised PolSAR image classification research is carried out with two basic steps: extracting features and training classifiers. The features of PolSAR data for traditional algorithms are mainly divided into four categories as follow:

1) Features based on Sinclair scattering matrix [S], such as copolar ratio, cross-polar ratio, copolar correlation coefficient and cross-polar correlation coefficient [1].
2) Polarimetric target decomposition features based on covariance matrix [C], coherence matrix [T] and [S], such as Pauli decomposition [1], Yamaguchi four-component decomposition [2], Freeman-Durden decomposition [3], and Huynen decomposition [4], etc.
3) Texture features, such as Gabor [5], LBP [6], SIFT [7], etc.
4) Spatial semantic features obtained from basic physical features, like the bag-of-word features based on Sparse Coding [8] and Spatial Pyramid Matching [9] algorithms.

In training step, classifiers like SVM [10], [11], CRF [12], Random Forest [13] and some ensemble classifiers [14], [15] are wildly implemented in PolSAR image classification studies.

These years, deep neural network algorithms have surpassed traditional optical image algorithms in many fields of computer vision [16], [17]. Recently, studies [18]-[20] have applied deep learning algorithms to PolSAR image classification tasks and achieved excellent results. Different from traditional methods, deep neural networks can adaptively capture features and learn classifiers from specific datasets and tasks, which are more flexible and robust. CNNs and RNNs are two widely-used deep neural networks, and CNNs have made a series of breakthroughs in image classification, target detection and semantic segmentation. For example, Microsoft research team outperforms human-level performance (5.1%) on ImageNet 2012 classification dataset [21], and Wu *et al*. proposed the CheXNet which can detect pneumonia from chest X-rays at a level outperforming experienced radiologists [22].

In [23]-[25], CNN, complex-valued CNN and graph-based CNN are used to implement precise PolSAR image classification pixel by pixel. However, they all have a large number of repeated calculations when extracting features from adjacent pixels, which waste a lot of time especially on high-resolution images. What's worse, graph-based CNN requires a lot of extra time spending on CRF model. The pixel mapping networks such as fully convolution neural network [26], dilated convolution neural network [27] and encoder-decoder network [28] can solve this problem by sharing convolutional parameters on whole images. For encoder-decoder network, the downsampled fe-

This work was supported in part by the National Natural Science Foundation of China under Grant 61471340, and the National Key Research and Development Program under Grant 2017YFB0503001.

The authors are with the Department of Airborne Microwave Remote Sensing System, Institute of Electrics, Chinese Academy of Sciences, Beijing, 100190 China (e-mail: xdluestc@outlook.com; cliu@mail.ie.ac.cn; ionlight@sina.com; wangchao_thu@163.com; xzhang@mail.ie.ac.cn;)

ature map from the encoder network is up-sampled to the original size by the decoder network, which achieves pixel-to-pixel mapping classification directly. In addition, the accuracy is improved for many contextual semantic features being captured. The pixel mapping classification networks include FCN [26], SegNet [28], DeepLab [29], and U-Net [30], etc. Moreover, pixel mapping networks have been applied in PolSAR image classification studies recently, such as a stacked FCN [26] based on scattering matrix and scattering coding feature, a graph embedded FCN [31] aiming at precise classification, and cascaded FCN (CasNet) [32] for road detection and centerline extraction tasks.

As for prior labels, pixel mapping networks require prior labeled maps corresponding to input images, which is difficult for PolSAR images because there are always some unlabeled pixels in their ground truth maps. So, in [33], Li *et al*. used a zero-initialized ground truth map with labeled training pixels in it to train the FCN for PolSAR classification, which is also adopted in [34]. However, this method leads to an inadequate trained network and edge-rough classification. In order to solve this problem, we propose a transfer dilated CNN to pre-classify input images, then merge the dense classified map with labels of training pixels to train the parallel pixel mapping network. In this way, sampled pixels are trained twice for higher accuracy.

Unlike optical images, the training data for PolSAR classification network is based on the complex-value covariance matrix [C] and coherence matrix [T] instead of the 3-channel RGB vector. The non-diagonal elements of the [C] and [T] matrices are complexes, and each pixel of the PolSAR images has rich polarimetric target decomposition features and channel phase information, which is helpful to improve the classification accuracy. In [24], complex-value convolution network is proposed to extract phase information, and all the network parameters are complexes, which is less portable. In [33], a scattering coding method is proposed to deal with complex scattering matrix [S], however this method expands input images by four times, which greatly increases the amount of the calculation.

Inspired by the previous studies [24][33][34] and the existing problems, in this paper, we propose a novel PolSAR image classification algorithm, i.e., a pixel-refining parallel mapping network in the complex domain (CRPM-Net) to take efficiency, accuracy and prior labels into consideration at the same time. The main contribution of this paper can be briefly summarized as follows.

1) We extend the proposed algorithm to complex domain by group convolution and cross-combination with complex cross-convolution kernel.
2) A transfer dilated convolution neural network in the complex domain (C-Dilated CNN) activated by complex cross-convolution neural network (Cs-CNN) is proposed to achieve precise localization and fast pixel mapping classification with a small number of prior labels.
3) We design a novel parallel pixel mapping network consists of C-Dilated CNN and complex-domain encoder-decode network to extract contextual semantic features and refine misclassified training pixels for higher accuracy, which is called CRPM-Net.
4) We train CRPM-Net with the fusion of dense score maps obtained from C-Dilated CNN and labels of high-weighted training pixels.

This paper is organized as follows, Section II presents the basic principles of CNN and Cs-CNN. Session III gives a detailed introduction to the structure and training framework of CRPM-Net. Experiments and results are described and analyzed in Session IV on four PolSAR datasets. And the conclusion is discussed in Section V.

## II. PRINCIPLES OF COMPLEX CROSS-CONVOLUTION NEURAL NETWORKS

This section briefly introduces the basic principles of convolutional neural networks (CNN) and complex cross-convolutional neural networks (Cs-CNN), as well as the gradient back propagation of Cs-CNN.

### A. Principles of CNN

CNN is inspired by the study of visual imaging [35], and it consists of three basic structures: convolution layer, activate layer, and pooling layer. The convolution layer captures sparse local features by sliding on the image with a 3-dimensional convolution kernel, and each convolution kernel is shared in the whole image. Output of the *j*-th convolution kernel $w^j$ on image *x* is calculated by

$$y_j = \sum_{i=1}^{c} \left[ \sum_{p+m, q+n \in S} x^i_{p+m, q+n} w^{ji}_{m,n} \right] + b_j \ , \ w^j \in \mathbb{R}^{k \times k \times c} \quad (1)$$

where *k* and *c* are the convolution kernel size and the number of image input channels, respectively. (*p*,*q*) is the upper left corner coordinate of the convolution kernel in image. $b_j$ is the bias corresponding to $w^j$, and *S* is the area covered by $w^j$. Then, $y_j$ is input into the active function for nonlinear representation, such as Sigmoid, Tanh, ReLU, CReLU [21], peaky ReLU [36], etc. Finally, the pooling layer downsamples the output of the active layer to increase the receptive field of each pixel, and the computational complexity of the next layer is reduced. The structure of a single layer CNN is shown in Fig. 1, The *0*-th and *1*-st convolution kernels slide on the input image *x* to obtain the linear feature map $y^0$, $y^1$, respectively. Through the active layer and pooling layer, $y^0$, $y^1$ are activated and downsampled to $z^0$, $z^1$.

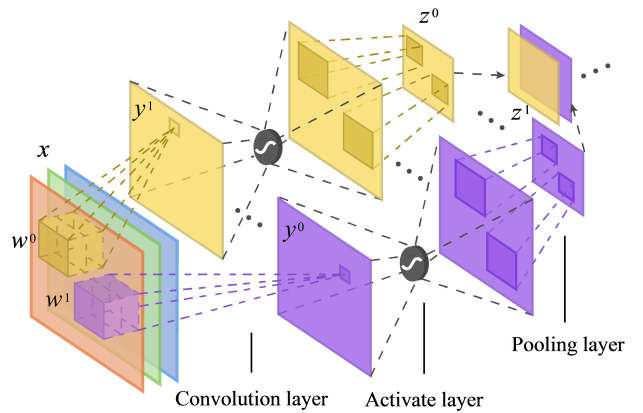

Fig. 1 Structure of single layer convolutional neural network

In multi-layer CNNs, the low-layer convolution extracts edge and corner features for good localization and the high-layer convolution extracts contextual semantic features. Those years, a lot of deep CNNs are designed to learn powerful features for classification, detection and segmentation tasks, such as VGG [37], GoogleNet [38], ResNet [39], and DenseNet [40], etc.

### B. Principles of Complex Cross-CNN

Since the phase difference between polarimetric channels can also be an important feature for PolSAR image classification, we extend the general CNN to the complex domain by using complex cross-convolution. In the complex domain, the input image $X$ consists of two real-value image groups representing the real part and imaginary part, respectively, and it can be defined as

$$X = \{x_0, x_1, ..., x_c\} = X_r + jX_i$$
$$= \{x_{0\_r}, x_{1\_r}, \cdots, x_{c\_r}\} + j\{x_{0\_i}, x_{1\_i}, \cdots, x_{c\_i}\} \quad (2)$$

$$X \in \mathbb{C}^{m \times n \times c}, x_c \in \mathbb{C}^{m \times n}, X_r, X_i \in \mathbb{R}^{m \times n \times c}, x_{c\_r}, x_{c\_i} \in \mathbb{R}^{m \times n}$$

where $c$ is the number of input channels, $X_r$, and $X_i$ represent the real and imaginary groups of the input image, respectively.

Besides, the convolution kernel is separated to the real group and imaginary group as well, which is expressed as

$$W = \{w_0, w_1, ..., w_o\} = W_r + jW_i \quad (3)$$

$$= \begin{bmatrix} w_{00\_r} & w_{01\_r} & \cdots & w_{0o\_r} \\ w_{10\_r} & w_{11\_r} & \cdots & w_{1o\_r} \\ \vdots & \vdots & \ddots & \vdots \\ w_{c0\_r} & w_{c1\_r} & \cdots & w_{co\_r} \end{bmatrix} + j \begin{bmatrix} w_{00\_i} & w_{01\_i} & \cdots & w_{0o\_i} \\ w_{10\_i} & w_{11\_i} & \cdots & w_{1o\_i} \\ \vdots & \vdots & \ddots & \vdots \\ w_{c0\_i} & w_{c1\_i} & \cdots & w_{co\_i} \end{bmatrix}$$

$$W \in \mathbb{C}^{c \times k \times k \times o}, w_o \in \mathbb{C}^{c \times k \times k}, W_r, W_i \in \mathbb{R}^{c \times k \times k \times o}, w_{co\_r}, w_{co\_i} \in \mathbb{R}^{k \times k}$$

where $o$ is the number of 3-dimention convolution kernels and is also the number of output channels.

Significantly, the real and imaginary groups of input images and convolutions have the same size and number of channels.

*1) Complex Cross-Convolution*: The general convolution operation is essentially the inner product of the 3-dimensional convolution kernel with the image window, and we define this operation as $Conv(\cdot)$. Moreover, the complex cross-convolution is a combination of four staggered convolution operations to achieve complex number multiplication, which is recorded as $\mathbb{C}onv(\cdot)$. Output of the $o$-th complex cross-convolution can be calculated by

$$y_o = \mathbb{C}onv(X, w_o) = \sum_{k=0}^{c} \mathbb{C}onv(x_k, w_{ko})$$
$$= \sum_{k=0}^{c} \mathbb{C}onv(x_{k\_r} + jx_{k\_i}, w_{ko\_r} + jw_{ko\_i})$$
$$= \sum_{k=0}^{c} \begin{bmatrix} Conv(x_{k\_r}, w_{ko\_r}) - Conv(x_{k\_i}, w_{ko\_i}) \\ + j(Conv(x_{k\_r}, w_{ko\_i}) + Conv(x_{k\_i}, w_{ko\_r})) \end{bmatrix} \quad (4)$$
$$= Conv(X_r, w_{o\_r}) - Conv(X_i, w_{o\_i})$$
$$\quad + j[Conv(X_r, w_{o\_i}) + Conv(X_i, w_{o\_r})]$$

where $w_{o\_r}, w_{o\_i}$ are the real part and imaginary part of the $o$-th convolution kernel, respectively. They are given by

$$\begin{cases} w_{o\_r} = \{w_{0o\_r}, w_{1o\_r}, \cdots, w_{co\_r}\}, & w_{o\_r} \in \mathbb{R}^{c \times k \times k} \\ w_{o\_i} = \{w_{0o\_i}, w_{1o\_i}, \cdots, w_{co\_i}\}, & w_{o\_i} \in \mathbb{R}^{c \times k \times k} \end{cases} \quad (5)$$

From (4), we can see that the output of the complex cross-convolution is split into four groups, which are convolution output of $X_r$ with $w_{o\_r}$, $X_i$ with $w_{o\_i}$, $X_r$ with $w_{o\_i}$, and $X_i$ with $w_{o\_r}$, respectively. The complex cross-convolution operation is illustrated in Fig. 2.

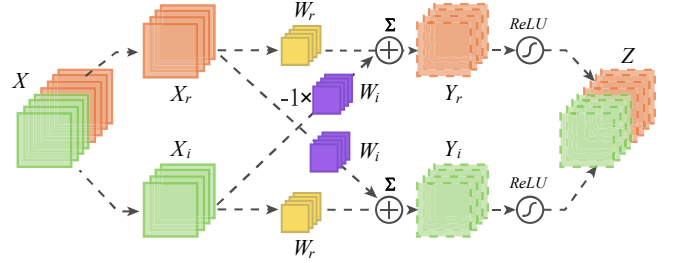

Fig. 2 Complex cross-convolution operation

According to Fig. 2, real and imaginary parts of the activation function are independent as well. The complex ReLU function used in this paper is given by

$$z_o = \delta(y_o) = \delta(y_{o\_r}) + j\delta(y_{o\_i}) = [y_{o\_r}]_+ + j[y_{o\_i}]_+$$
$$[y]_+ = \begin{cases} y & y > 0 \\ 0 & else \end{cases} \quad (6)$$

*2) Gradient Back Propagation*: In the training process of deep learning, variables are updated with negative gradients of the loss. As for a single layer complex cross-convolution neural network (Cs-CNN), the loss gradient of $w_{ko}$ is written as

$$\frac{\partial L(z_o)}{\partial w_{ko}} = \frac{\partial L(z_o)}{\partial w_{ko\_r}} + j\left[\frac{\partial L(z_o)}{\partial w_{ko\_i}}\right], \quad L(z_o) \in \mathbb{R} \quad (7)$$

The imaginary part of the loss gradient can be expanded as

$$\frac{\partial L(z_o)}{\partial w_{ko\_i}} = \frac{\partial L(z_o)}{\partial z_{o\_r}} \frac{\partial z_{o\_r}}{\partial y_{o\_r}} \frac{\partial y_{o\_r}}{\partial w_{ko\_i}} + \frac{\partial L(z_o)}{\partial z_{o\_i}} \frac{\partial z_{o\_i}}{\partial y_{o\_i}} \frac{\partial y_{o\_i}}{\partial w_{ko\_i}}$$
$$= \delta'(y_{o\_i}) \frac{\partial L(z_o)}{\partial z_{o\_i}} \frac{\partial Conv(x_{k\_r}, w_{ko\_i})}{\partial w_{ko\_i}} \quad (8)$$
$$- \delta'(y_{o\_r}) \frac{\partial L(z_o)}{\partial z_{o\_r}} \frac{\partial Conv(x_{k\_i}, w_{ko\_i})}{\partial w_{ko\_i}}$$

The real part of the loss gradient can be expanded as

$$\frac{\partial L(z_o)}{\partial w_{ko\_r}} = \frac{\partial L(z_o)}{\partial z_{o\_r}} \frac{\partial z_{o\_r}}{\partial y_{o\_r}} \frac{\partial y_{o\_r}}{\partial w_{ko\_r}} + \frac{\partial L(z_o)}{\partial z_{o\_i}} \frac{\partial z_{o\_i}}{\partial y_{o\_i}} \frac{\partial y_{o\_i}}{\partial w_{ko\_r}}$$
$$= \delta'(y_{o\_r}) \frac{\partial L(z_o)}{\partial z_{o\_r}} \frac{\partial Conv(x_{k\_r}, w_{ko\_r})}{\partial w_{ko\_r}} \quad (9)$$
$$+ \delta'(y_{o\_i}) \frac{\partial L(z_o)}{\partial z_{o\_i}} \frac{\partial Conv(x_{k\_i}, w_{ko\_r})}{\partial w_{ko\_r}}$$

Substituting (8) and (9) to (7), the negative gradient of $w_{ko}$ is expressed as

$$-\frac{\partial L(z_o)}{\partial w_{ko}} = -\delta'(y_{o\_r})\frac{\partial L(z_o)}{\partial z_{o\_r}}\frac{\partial Conv(x_{k\_r},w_{ko\_r})}{\partial w_{ko\_r}} \quad (10)$$

$$-\delta'(y_{o\_i})\frac{\partial L(z_o)}{\partial z_{o\_i}}\frac{\partial Conv(x_{k\_i},w_{ko\_r})}{\partial w_{ko\_r}}$$

$$+j\left[\begin{array}{c}\delta'(y_{o\_r})\frac{\partial L(z_o)}{\partial z_{o\_r}}\frac{\partial Conv(x_{k\_i},w_{ko\_i})}{\partial w_{ko\_i}}\\-\delta'(y_{o\_i})\frac{\partial L(z_o)}{\partial z_{o\_i}}\frac{\partial Conv(x_{k\_r},w_{ko\_i})}{\partial w_{ko\_i}}\end{array}\right]$$

After t iterations, $w^t_{ko}$ is updated as (11), where $\eta$ is the gradient update ratio

$$w^t_{ko} = w^{t-1}_{ko} - \eta\frac{\partial L(z_o)}{\partial w_{ko}} \quad (11)$$

According to equation (10), the real and imaginary groups of complex cross-convolution kernel are updated separately.

## III. PROPOSED ALGORITHMS AND TRAINING FRAMEWORK

In this section, we introduce the proposed model and corresponding training algorithm in detail. And it is worth mentioning that we extend the model to the complex domain by complex cross-convolution mentioned in previous section so as to make full use of the phase information.

### A. Structure of Complex Cross-Convolution Neural Network

A 3-layer down-sampling complex cross-convolution neural network (Cs-CNN) is designed to extract polarimetric target decomposition features and phase information, as shown in Fig. 3(a). The filter sizes of the four cross-convolutional layers are 3×3, 3×3, 1×1, 1×1, respectively, and at each convolution layer except the last one we double the number of feature channels. The pooling layer is a 2×2 max-pooling with stride 2. In addition, there is no padding in the convolution so as to transfer network parameters to the C-Dilated CNN which will be described in the next chapter, so the image size will be reduced by 2 pixels in each 3×3 cross-convolution layer. Finally, a 1×1 cross-convolution is used to map each 48-dimension feature vector to the number of classes. The input feature window for each training pixels has a size of 10×10, and finally, a 1×1 complex-value output feature map is obtained.

In order to calculate loss with real-value labels and the 1×1 complex-value output feature map, we stack the real and imaginary parts of the feature map with its amplitude and phase to get a 4-channel feature map. Then, a 4×1 full connected layer is used to merge the stacked feature map to obtain score map for each class, which is shown in Fig. 4(a).

### B. Transfer Dilated Convolutional Neural Network in the complex domain

Since the Cs-CNN has a lot of redundant calculations when extracting features from adjacent pixels, it's inefficient to deal with whole images, especially high-resolution PolSAR images. Aim to improve the classification speed of Cs-CNN, we transfer its parameters to a dilated convolution neural network in the complex domain (C-Dilated CNN) with the same scale of parameters to implement dense pixel mapping of input and output images. In this way, images can be classified without being downsampled, so that the whole image classification speed is greatly accelerated. The structure of C-Dilated CNN is shown in Fig. 3(b).

In this paper, the C-Dilated CNN uses a set of 1×1 hole padding cross-convolution kernels and 2×2 max-pooling layers with stride 1 to obtain the same spatial receptive field per pixel as that of Cs-CNN, and the size of input image is unchanged. So, the C-Dilated CNN has much faster speed and similar classification capabilities compared with Cs-CNN. The red dotted line in Fig. 3 shows the transference of network parameters from the first and second layers of Cs-CNN to C-Dilated CNN.

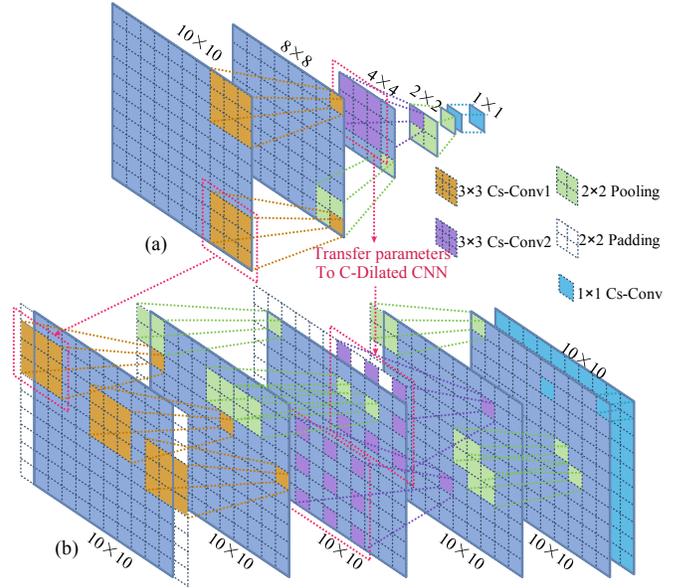

Fig.3 Structure of Cs-CNN and C-Dilated CNN. The differences between two networks are the dilated convolution kernel in 2-nd layer and one pixel pooling stride.

### C. Encoder-Decoder Network in the complex domain

Similar to C-Dilated CNN, the encoder-decoder network in the complex domain can also achieve efficient pixel mapping classification in consideration of phase information. This network's structure is illustrated in Fig.4(c).

In the PolSAR image classification task, the contextual features of the target are relatively weak except some sparse artificial terrains such as roads and buildings. Therefore, the encoder-decoder network is not required to capture the contextual semantic features with a large receptive field. So, the designed encoder network only contains three trainable complex cross-convolutional layers, which is the same as Cs-CNN, and its output is a 4-time downsampled feature map with contextual semantic features. In order to achieve pixel mapping, the decoder network up-samples the downsampled feature map to the original size by two transposition convolution layers [41], which is a learnable convolution sliding on input image padded at the edge or inside. Finally, we use the 1×1 cross-convolution to map each 48-dimension feature vector to the desired number of classes.

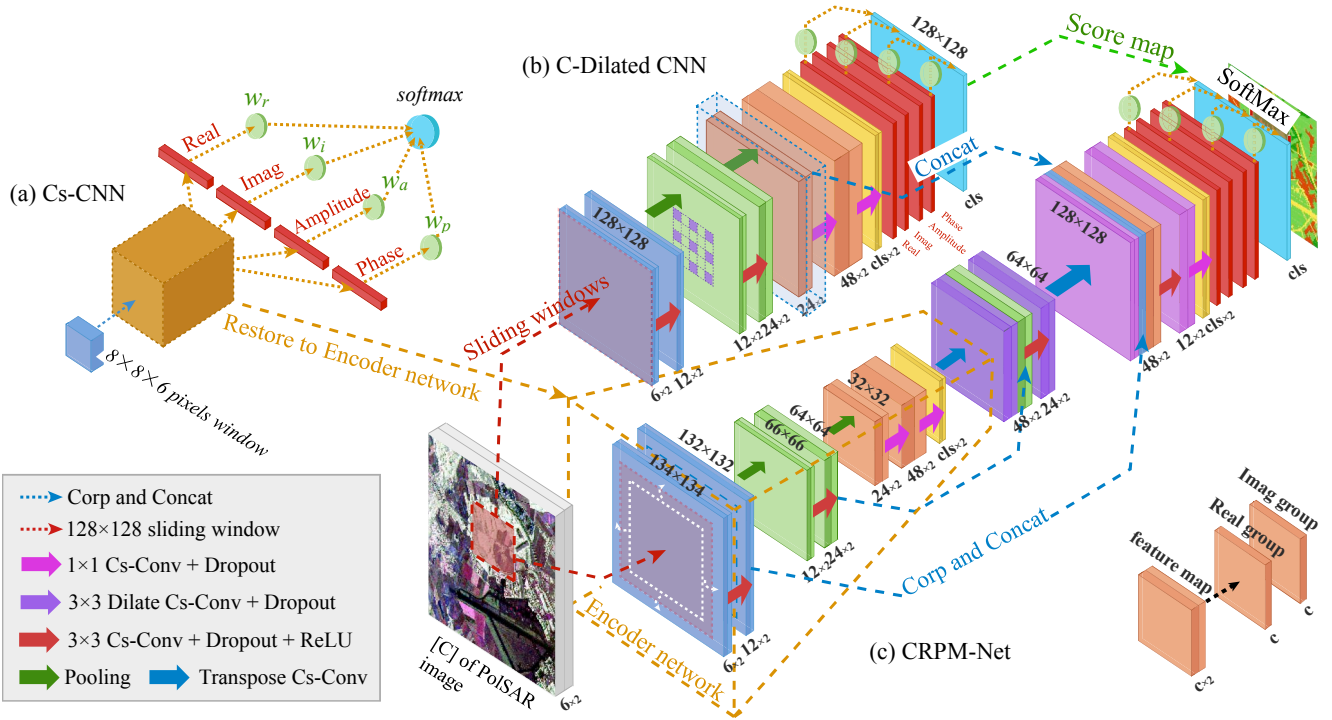

Fig.4 Structure of CRPM with Cs-CNN and C-Dilated CNN. (a) Cs-CNN. (b) C-Dilated CNN. (c) CRPM-Net. Red dotted lines represent the 128×128 sliding windows for training, blue dotted lines are corp-concat processes, orange dotted lines mean the Encoding network, and green dotted line is the dense score maps obtained by C-Dilated CNN, which are part of the training labels. The sizes of input images of C-Dilated CNN and encoder-decoder network are 128×128 and 134×134, respectively. The missing border pixels of 134×134 input images are extrapolated by mirroring shown in white dotted block in (c).

Since the decoder network loses a lot of terrain local information, the classification result is rough on the edges and corners. In order to localize, the high-resolution feature maps from the encoder network are combined with the upsampled output of the decoder network on the channel dimension. In this way, precise localization and the use of contextual semantic features are feasible at the same time. Before combining with feature maps of the decoder network, feature maps obtained from encoding network should be cropped for the loss of border pixels caused by unpadded convolutions as mentioned in Cs-CNN.

D. *Structure of CRPM-Net and Training Framework*

*1) Structure of complex domain pixel-refining parallel mapping network (CRPM-Net)*: In order to take both localization accuracy and contextual semantic features into account, we parallelly merge the C-Dilated CNN into an encoder-decoder network by concatenation with the 24-dimension feature map of C-Dilated CNN and the 24-dimension upsampled feature map of decoder network, which is illustrated in Fig. 4(c). However, C-Dilated CNN doesn't change the size of input images, whereas, encoder-decoder network will lose six border pixels for unpadded convolution. So, in order to obtain a final feature map with the same size as that of C-Dilated CNN, we extrapolate the missing border pixels by mirroring the input image for encoder-decoder network [as highlighted by white dotted box in Fig.4(c)]. In training process, a 128×128 window is adapted to sample on input images with stride 64, and the sampled image will be expanded to 134×134 for encoder-decoder network.

*2) Training Framework*: Both the C-Dilated CNN and encoder-decoder network require label maps corresponding to input images for training, which is quite dependent on full image ground truth. However, the full-labeled PolSAR images under the same imaging conditions such as band, projection direction and viewing angle are very insufficient, so it is difficult to train the CRPM–Net with a large number of identically distributed PolSAR images. To solve this problem, we propose a two-step training algorithm, firstly, we use the Cs-CNN trained with a small number of labeled pixels to activate the C-Dilated CNN and the encoder network, which can achieve fast pixel mapping classification. Secondly, the CRPM-Net is trained to extract contextual semantic features and refine the misclassified pixels for Cs-CNN and C-Dilated CNN. The two training steps can make full use of the limited labeled pixels to achieve precise and efficient classification.

In experiments, it is more difficult to distinguish the class with smaller number of training pixels. In response to this problem, we use the focal loss in Retina-Net [42] to reduce the loss of easily classified pixels, so that the training of the network pays more attention to the classes with fewer pixels or large training errors. Assuming the complex-value output of Cs-CNN is $z' = \lambda e^{j\varphi}$, then the focal loss is represented by the following formula

$$p(z') = Softmax\left(w_r \Re(z') + w_i \Im(z') + w_m \lambda + w_p \varphi\right) \quad (13)$$

$$L_{focal}(z', z) = -\alpha\left(1 - p_z(z')\right)^\gamma \log\left(p_z(z')\right) \quad (14)$$

where $z$ represents the true label of the current pixel and $p_z(z') \in \mathbb{R}^{class\_num}$ is the probability of it for class $z$.

**Algorithm 1** The training framework of CRPM-Net

**Input:** The filtered covariance matrix [C] of PolSAR image $C$, ground truth $G$, low sampling rate $r$;

1: Randomly sample training pixels on $G$, and the number of training pixels per class is similar, in addition, all sampling rates are less than $r$;
2: Train the 3-layer Cs-CNN by 10×10 feature areas centered on training pixels, and the focal loss is used in this step with $\alpha$=0.25, $\gamma$=2, learning rate $\eta_1$;
3: Transfer the parameters of Cs-CNN to C-Dilated CNN and the encoder network.
4: Merge C-Dilated CNN parallelly into encoder-decoder network with shared parameters, which is so-called CRPM-Net;
5: Slide 128×128 window on $C$ with stride 64 to get complex-value sub-images, recorded as $I_c$,
6: Put $I_c$ into C-Dilated CNN to get the dense classified map, recorded as $O$.
7: Get the refined dense score map $M$ and pixels loss weight matrix $W$ from **Algorithm 2**;
8: Train the decoder network of CRPM-Net, by using $I_c$ and $M$ as input, calculate the classification loss with $W$. The cross-entropy loss is adopted in this step, with learning rate $\eta_2$

**Ouput:** Learnt parameters of CRPM-Net.

**Algorithm 2** Refined score map and Loss weight matrix

**Input:** Training pixels $p$, score map of C-Dilated CNN $O$.

1: Initialize a weight matrix with the same size as $O$, recorded as $W$, and all the elements are assigned a value of $w_{else}$;
2: **if** $p$ in $O$ **then**
3:    Replace the corresponding pixel label in $O$ with label of $p$;
4:    **if** $p$ is misclassified **then**
5:      Replace the pixel weight in $W$ with the highest loss weight $w_{error}$;
6:    **else if** $p$ is classified correctly **then**
7:      Replace the pixel weight in $W$ with a high loss weight $w_{train}$;
8:    **end if**
9: **end if**

**Output:** Refined dense score map $M$ and pixels loss weight matrix $W$.

The training framework of CRPM-Net is introduced in detail in Algorithm 1. Considering the balance training for each class, we sample a similar number of pixels per class, and the sample rates of each class are all at a low ratio. 10×10 feature areas centered on each training pixel are used to pre-train the Cs-CNN with focal loss function and Adam optimizer [43]. After that, we transfer network parameters of Cs-CNN to C-Dilated CNN, which can achieve pixel mapping directly with just a small drop in accuracy. Then, Cs-CNN is restored to the encoder network of CRPM-Net. Finally, we train the decoder network of CRPM-Net based on the dense score map obtained from C-Dilated CNN and high-weighted training pixels to capture contextual semantic features and refine the misclassified pixels of Cs-CNN and C-Dilated CNN.

In the 7-th step of Algorithm 1, we restore the label of training pixels to the dense score map obtained from C-Dilated CNN, and form a loss weight matrix with different weights for training, misclassified, and residual pixels, such as $w_{train}$=10.0, $w_{error}$=50.0, $w_{else}$=1.0. Then, the new score map and loss weight matrix are used for training CRPM-Net with cross-entropy loss function and Adam optimizer. This part is described in detail in Algorithm 2.

Similar to Cs-CNN and C-Dilated CNN, the final complex-value feature map of CPRM-Net and its corresponding amplitude and phase are connected on the channel dimension. Then the connected real-value feature map is input into a full connected layer with the shape 4×1 to get the real-value classification score map. Finally, the cross-entropy loss is calculated to update parameters of decoder network.

## IV. EXPERIMENTS AND RESULTS ANALYSIS

To verify the classification ability of the proposed algorithm in this paper, we compare the general convolution neural network (CNN), complex cross-convolution neural network (Cs-CNN), real and complex domain dilated convolution network (Dilated CNN, C-Dilated CNN) and real and complex domain pixel-refining parallel mapping network (RPM-Net, CRPM-Net) on four PolSAR images from different platforms. At last, we cross compare the CRPM-Net with two recent state-of-the art studies to further point out the competitive performance of the proposed CRPM-Net.

The training data is based on the covariance matrix [C] after speckle filtering by refined Lee algorithm [44]. In real domain networks, the dimension of feature vector referred to (15) is 9 in the condition of reciprocal backscattering $S_{HV} = S_{VH}$ [20], where $C_{11}$, $C_{22}$, $C_{33}$ are the diagonal elements of the [C] matrix respectively, and $C_{12}$, $C_{13}$, $C_{23}$ are the non-diagonal complex elements of the [C] matrix respectively.

$$features_{real} = \begin{bmatrix} C_{11}, \Re(C_{12}), \Im(C_{12}), \\ C_{22}, \Re(C_{13}), \Im(C_{13}), \\ C_{33}, \Re(C_{23}), \Im(C_{23}) \end{bmatrix} \quad (15)$$

For complex domain networks, the pixel of input image is a 6-dimensional complex-value feature vector: $features_{imag}$ = [ $C'_{11}$, $C'_{22}$, $C'_{33}$, $C_{12}$, $C_{13}$, $C_{23}$], where the imaginary parts of $C_{11}$, $C_{22}$, $C_{33}$ are $10^{-8}$.

Besides, the Z-score normalization is required for each feature dimension as a preprocessing method, which is defined by:

$$\hat{C} = \frac{C - \frac{1}{N}\sum_{imag} C^{ij}}{\sqrt{\frac{1}{N}\sum_{imag}\left(C^{ij} - \frac{1}{N}\sum_{imag} C^{ij}\right)^2}} \quad (16)$$

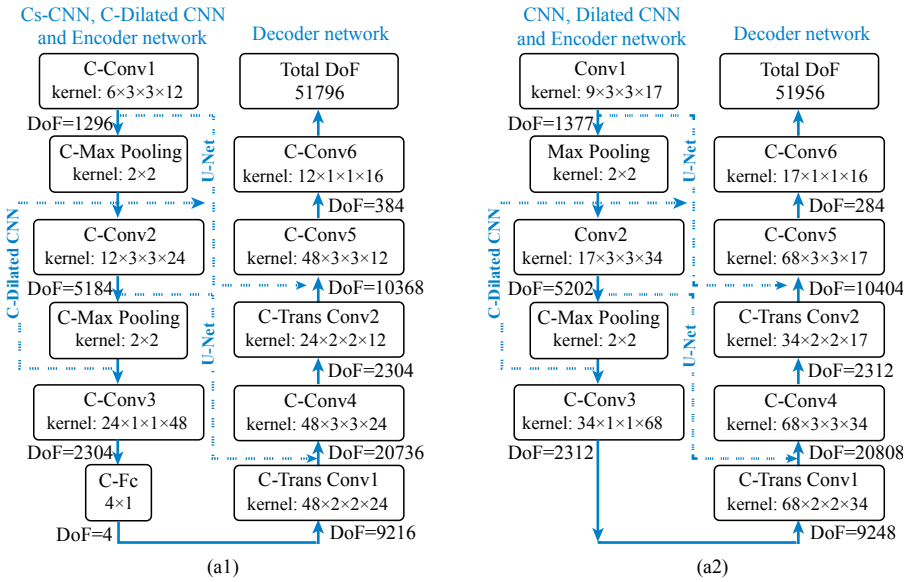
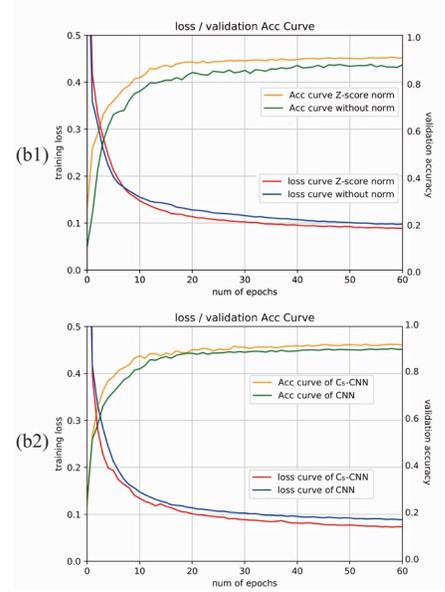

Fig.5 Parameters and convergence curve of complex-domain networks and real-domain networks. (a1) CRPM-Net (a2) RPM-Net. For the first experiment: (b1) convergence curve of CNN. (b2) convergence curve of CNN and Cs-CNN.

Experiments show that networks converge faster and achieve smaller losses and higher verification accuracy with Z-score normalization. The convergence curves of Z-score normalization and none normalization are shown in Fig. 5(b1).

In order to compare real and complex domain networks fairly, we set similar parameters for these two types of networks. The output channel dimension of the 1st convolution layer in Cs-CNN is 12, and the freedom of degree (DoF) of parameters of whole complex domain network is 51796, as shown in Fig. 5(a1). Besides, the output channel dimension of the 1-st convolution layer in CNN is 17, and the DoF of whole real domain network is 51956, as shown in Fig. 5(a2).

The performances of different algorithms are compared by the confusion matrix $N \in \mathbb{R}^{cls \times cls}$, overall accuracy($OA$), $FWIoU$, $Kappa$ coefficient and classification time. Among them, The $FWIoU$ is defined by

$$FWIoU = \frac{1}{\sum_{i=0}^{cls}\sum_{j=0}^{cls} N_{ij}} \sum_{i=0}^{cls} \left( \frac{N_{ij}}{\sum_{i=0}^{cls} N_{ij} + \sum_{j=0}^{cls} N_{ji} - N_{ii}} \right) \quad (17)$$

And the $Kappa$ coefficient is calculated by

$$Kappa = \frac{\sum_{i=1}^{cls} N_{ii} N_{total} - \sum_{i=1}^{cls} \left( \sum_{j=1}^{cls} N_{ij} \sum_{j=1}^{cls} N_{ji} \right)}{(N_{total})^2 - \sum_{i=1}^{cls} \left( \sum_{j=1}^{cls} N_{ij} \sum_{j=1}^{cls} N_{ji} \right)} \quad (18)$$

All networks are trained and evaluated based on Tensorflow deep learning framework [45], NVIDIA GeForce GTX 1070-ti GPU (8 GB) and Intel Xeon CPU (2.67GHz).

The learning rate $\eta_1$, batch size, and max training epoch of CNN, Cs-CNN training step are 0.005, 100 and 60 respectively for all experiments. Additionally, the learning rate $\eta_2$, batch size, and max training epoch of RPM-Net, CRPM-Net training step are 0.001, 5 and 30 respectively for all experiments. Especially, the loss weight matrix $W$ is valued by $w_{train}=50.0$, $w_{error}=100.0$, $w_{else}=0.5$ in all four experiments.

### A. Experiment on Flevoland-Netherlands, AIRSAR, L,P,C-Band PolSAR Dataset

The first experiment is carried out on a *L,P,C*-band full PolSAR image over the Flevoland-Netherlands region. It is acquired by the NASA/Jet Propulsion Laboratory AIRSAR platform [46], which is usually used as a benchmark data for PolSAR classification research. This PolSAR image data contains a variety of crops and artificial targets, and the pseudo RGB image synthesized by its *L,P,C*-band *SPAN*s is shown in Fig. 6(a). Image size is 1279×1024 pixels. There are 16 classes in total including potatoes, beet, peas, barley, beans, wheat, flax, fruits, lucerne, oats, rapeseed, maize, onions, roads and some buildings. The ground truth and label legends are shown in Fig. 6(b) and (c) [47], where the black pixels are not involved in the experiment. We selected training samples for each class with a similar number in ground truth area, the number of samples for each class is shown in the second column of Tab. I. Except peas, the sampling rate of each class is less than 6%, and in [23], 10% is verified as a suitable sampling rate for this PolSAR image data, so we choose a lower sampling rate to verify the generalization ability of our algorithm.

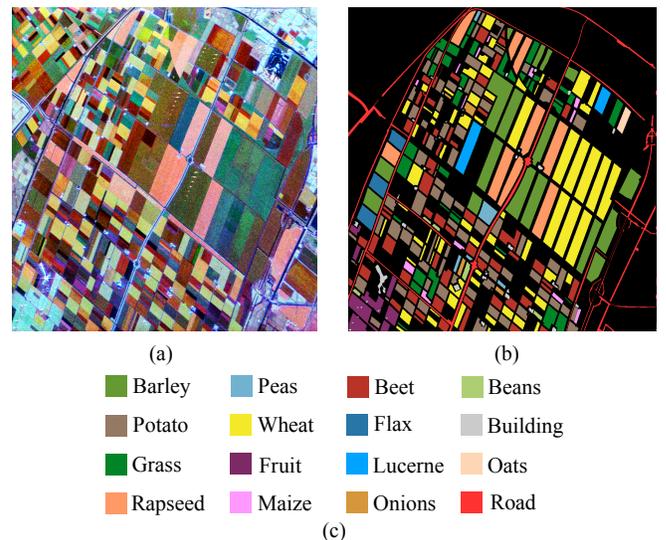

Fig. 6 *L,P,C*-band full PolSAR image data over the Flevoland-Netherlands region. (a) Pseudo RGB image. (b) Ground truth. (c) Label legends.

TABLE I
CLASSIFICATION RESULTS OF FLEVOLAND-NETHERLANDS POLSAR IMAGE

| Class (%) | Train | Total | SVM | CNN | Dilated CNN | RPM-Net | Cs-CNN | C-Dilated CNN | CRPM-Net |
|---|---|---|---|---|---|---|---|---|---|
| Grass (2.0) | 600 | 29630 | 92.53 | 96.04 | 96.20 | **97.32** | 96.52 | 96.52 | 96.60 |
| Flax (4.9) | 400 | 8031 | 98.39 | 99.68 | 99.76 | **99.87** | 99.55 | 99.65 | 99.73 |
| Potatoes (0.9) | 600 | 66148 | 97.10 | 98.84 | 98.63 | 98.85 | 98.93 | 98.86 | **99.13** |
| Wheat (0.6) | 600 | 90681 | 99.01 | 99.09 | 99.16 | 99.08 | 99.15 | 99.18 | **99.35** |
| Rapeseed (1.6) | 600 | 38015 | 95.45 | **100.00** | 99.98 | 99.99 | 99.99 | 99.97 | **100.00** |
| Beet (1.5) | 700 | 46583 | 91.26 | 94.89 | 94.68 | 95.17 | 98.28 | 98.43 | **98.79** |
| Barley (1.0) | 700 | 68427 | 98.12 | 97.99 | 97.85 | 98.81 | 99.33 | 99.28 | **99.44** |
| Peas (7.2) | 300 | 4155 | 96.36 | 92.90 | 93.02 | 91.52 | 94.66 | 91.70 | **98.42** |
| Maize (5.2) | 240 | 4611 | 89.22 | 86.05 | 82.88 | 92.19 | 92.37 | 90.63 | **95.63** |
| Bean (5.0) | 100 | 1982 | 96.09 | 96.47 | 95.66 | 96.16 | **97.78** | 97.26 | 97.18 |
| Fruit (4.4) | 600 | 13485 | 99.05 | 98.82 | 98.93 | **99.08** | 98.06 | 97.96 | 98.12 |
| Onion (4.5) | 100 | 2206 | 92.38 | 97.64 | 96.37 | 98.54 | **98.96** | 98.23 | 98.81 |
| Oats (5.4) | 100 | 1838 | 99.67 | 100.00 | 100.00 | 100.00 | 100.00 | 100.00 | 100.00 |
| Lucerne (4.9) | 400 | 8203 | 98.25 | 99.86 | 99.86 | **100.00** | 99.80 | 99.99 | 99.56 |
| Buildings (4.4) | 200 | 4536 | 26.15 | 82.08 | 75.28 | 81.51 | **87.52** | 81.16 | 84.51 |
| Road (1.1) | 400 | 42861 | 54.48 | 63.12 | 57.19 | 60.54 | 68.03 | 62.70 | **70.96** |
| **OA** | \ | \ | 92.44 | 95.13 | 94.51 | 95.26 | 96.26 | 95.62 | **96.63** |
| **Kappa** | \ | \ | 91.10 | 94.32 | 93.57 | 94.46 | 95.65 | 94.89 | **96.08** |
| **FWIoU** | \ | \ | 85.66 | 91.17 | 90.30 | 91.36 | 93.36 | 92.27 | **93.94** |
| **Pred Time** | \ | \ | 669 s | 220.31s | **3.94 s** | 4.12 s | 401.56 s | 4.32 s | 4.68 s |

The input channels of real and complex domain networks are 27 and 18 respectively, and convergence curves of Cs-CNN and CNN are illustrated in Fig. 5(b2), which shows that Cs-CNN converges faster and more stably than CNN and obtains better classification results finally since Cs-CNN extracts rich polarimatric target decomposition features and phase information.

From the two columns of CNN and Cs-CNN in Tab. I, we can see that the *OA*, *Kappa* and *FWIoU* of Cs-CNN are about 1.1 percentage points higher than CNN's. Furthermore, the accuracy of each class with fewer pixels such as maize, peas, beans, and onions is 1.5 percentage points higher. However, the full image classification time of Cs-CNN is about 2 times more than that of CNN. After the direct parameters transference from Cs-CNN, *OA* of C-Dilated CNN dropped by 0.63 percentage points, but the classification speed of C-Dilated is more than 90 times faster than that of CNN.

TABLE II
CONFUSION MATRIX OF CRPM-NET MODEL ON FLEVOLAND-NETHERLANDS IMAGE

| % | 1 | 2 | 3 | 4 | 5 | 6 | 7 | 8 | 9 | 10 | 11 | 12 | 13 | 14 | 15 | 16 |
|---|---|---|---|---|---|---|---|---|---|---|---|---|---|---|---|---|
| 1 Grass | **96.6** | 0 | 0 | 0.1 | 0 | 0 | 0.2 | 0 | 0.2 | 0 | 0 | 0.4 | 0 | 0 | 0.4 | 2.0 |
| 2 Flax | 0.1 | **99.7** | 0 | 0 | 0 | 0 | 0 | 0 | 0 | 0 | 0 | 0 | 0 | 0 | 0.2 | 0 |
| 3 Potatoes | 0.1 | 0 | **99.1** | 0 | 0 | 0.1 | 0 | 0 | 0.2 | 0 | 0 | 0 | 0 | 0 | 0.3 | 0.2 |
| 4 Wheat | 0.2 | 0 | 0 | **99.4** | 0 | 0 | 0.2 | 0 | 0 | 0 | 0 | 0 | 0 | 0 | 0 | 0.2 |
| 5 Rapeseed | 0 | 0 | 0 | 0 | **100** | 0 | 0 | 0 | 0 | 0 | 0 | 0 | 0 | 0 | 0 | 0 |
| 6 Beet | 0.2 | 0 | 0.1 | 0 | 0 | **98.8** | 0 | 0.2 | 0.2 | 0 | 0 | 0.4 | 0 | 0 | 0 | 0.1 |
| 7 Barley | 0.4 | 0 | 0 | 0 | 0 | 0 | **99.4** | 0 | 0 | 0 | 0 | 0 | 0 | 0 | 0.2 | 0 |
| 8 Peas | 0.3 | 0 | 0 | 0 | 0 | 0 | 0 | **98.4** | 1.1 | 0 | 0 | 0 | 0 | 0 | 0 | 0.2 |
| 9 Maize | 1.9 | 0 | 0.1 | 0 | 0 | 1.8 | 0 | 0 | **95.6** | 0.2 | 0 | 0 | 0 | 0 | 0 | 0.4 |
| 10 Bean | 1.1 | 0 | 0 | 0 | 0 | 0 | 0 | 0 | 0 | **97.2** | 0 | 0 | 0 | 0 | 0 | 1.7 |
| 11 Fruit | 0.2 | 0 | 0.1 | 0 | 0 | 0.1 | 0 | 0 | 0 | 0 | **98.1** | 0 | 0 | 0 | 1.5 | 0 |
| 12 Onion | 0 | 0 | 0 | 0 | 0 | 1.0 | 0 | 0.2 | 0 | 0 | 0 | **98.8** | 0 | 0 | 0 | 0 |
| 13 Oats | 0 | 0 | 0 | 0 | 0 | 0 | 0 | 0 | 0 | 0 | 0 | 0 | **100** | 0 | 0 | 0 |
| 14 Lucerne | 0.1 | 0 | 0 | 0 | 0 | 0 | 0 | 0.3 | 0 | 0 | 0 | 0 | 0 | **99.6** | 0 | 0 |
| 15 Buildings | 1.2 | 0 | 1.1 | 0 | 0.7 | 1.4 | 0 | 0 | 0 | 0 | 1.6 | 0 | 0 | 0 | **84.5** | 9.4 |
| 16 Roads | 20.9 | 0.1 | 1,7 | 1.1 | 1.3 | 1.6 | 0.5 | 0.2 | 0.3 | 0.2 | 0.1 | 0 | 0 | 0 | 1.0 | **71.0** |

*Note: $N_{ij}$ of confusion matrix means the *i*-th class is classified to the *j*-th class. Each row represents the true class, each col represents the class predicted;

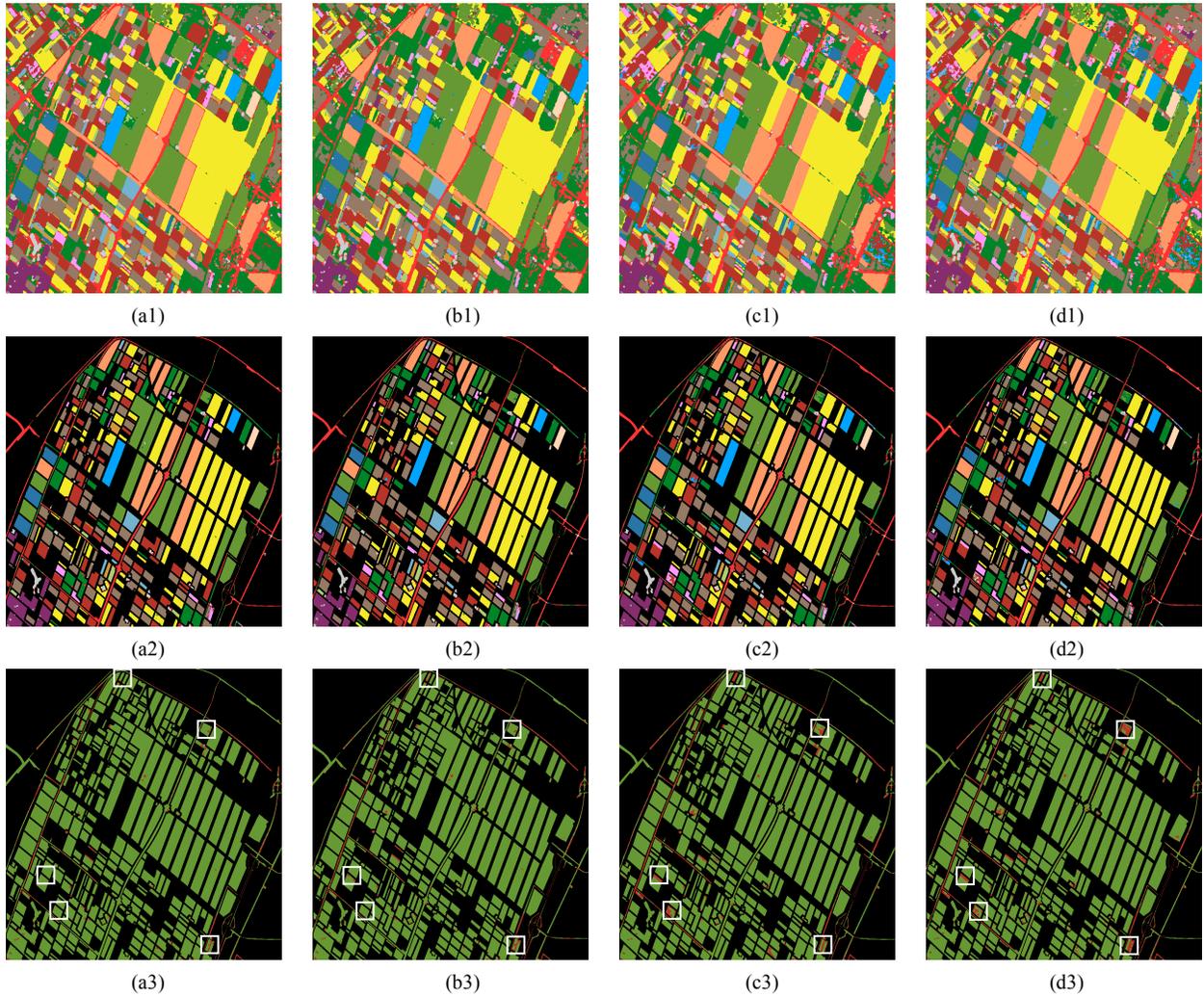

Fig. 7 Classification images of the Flevoland-Netherlands region. (a-d): CRPM-Net, C-CNN, RPM-Net, CNN. (1-3): classification results of full image, ground truth areas and error maps. As for error maps, the green and red areas are correctly classified and misclassified pixels in ground truth. The main differences between the results of the six models are shown in white rectangles in error maps

Finally, CRPM-Net achieves the best classification accuracy in potatoes, wheat, beet, barley, peas, maize, oats and roads, and the results of *OA*, *Kappa* and *FWIoU* are 96.63%, 96.08%, 93.94%, respectively, which are also the highest compared with other models according to the last column in Tab. I. As for the time cost, CRPM-Net spends 4.82 seconds on whole image classification which is about 83 times less than that of Cs-CNN, and 0.6 seconds more than that of C-Dilated CNN. The results indicate the excellent performance of CRPM-Net on the PolSAR image over Flevoland-Netherlands region.

The classification confusion matrix of CRPM-Net is shown in Tab. II. From the diagonal elements, we can see that the number of each terrain class varies greatly, however, classes such as beans, maize, onions, peas which have small number of training pixels, are well classified as well. On the other hand, roads are most confusing in this experiment, since roads rely on more edge information and are easily confused with other classes on the boundary. Moreover, many maize pixels are misclassified into beet since training pixels of maize is much less than beet.

The classification results are all in good agreement with the ground truth at the first glance as shown in Fig. 7. Furthermore, CRPM-Net can reduce the number of holes and discrete pixels regions. In a word, CRPM-Net can achieve the fairly fast and most accurate classification on PolSAR image over Flevoland-Netherlands with a small number of training pixels.

### B. Experiment on Oberpfaffenhofen, E-SAR, L-Band Dataset

In order to verify the robustness of the proposed algorithm for terrains with irregular surface such as urban, wood and mountains, we choose the *L*-band full PolSAR image over the Oberpfaffenhofen region for experiment, which is acquired by the E-SAR platform [46]. There are large rough urban and wood areas in this image, its Pauli RGB image is shown in Fig. 8(a). The size of the image is 1300×1200 pixels, the ground truth and label legend are shown in Fig. 8(b) and (c) respectively, which is acquired manually according to remote-sensing imagery in Google Earth over the Oberpfaffenhofen region. There are four classes in total: urban, farmland, roads and woodland, while black pixels in ground truth are not involved in the experiment. The number of training pixels and total pixels for each class are shown in the second and third columns in Tab. III. In addition, the sampling rate of each class is less than 2%.

The dimension of input feature vectors for real and complex domain networks is 9 and 6 respectively, since only L-Band is available. According to Tab. III, Cs-CNN is successful for urban and woodland with the *OA* of 89.71%, the *Kappa* of 0.8279 and the *FWIoU* of 0.8171, whereas, its classification time is 512.43 seconds.

TABLE III
CLASSIFICATION RESULTS OF OBERPFAFFENHOFEN POLSAR IMAGE

| Class (%) | Train | Total | CNN | Dilated CNN | RPM-Net | Cs-CNN | C-Dilated CNN | CRPM-Net |
|---|---|---|---|---|---|---|---|---|
| WoodLand(1.0) | 3000 | 290915 | 88.67 | 88.48 | 90.47 | 90.70 | 89.93 | **91.76** |
| FarmLand(0.5) | 3000 | 622739 | 93.82 | 94.10 | **95.00** | 93.31 | 94.12 | 94.60 |
| Urban (1.1) | 3000 | 274684 | 82.96 | 80.01 | 85.92 | 85.98 | 81.94 | **89.43** |
| Roads (1.8) | 2500 | 137400 | 79.46 | 76.29 | 77.79 | **78.62** | 75.85 | 76.66 |
| *OA* | \ | \ | 88.97 | 88.12 | 90.34 | 89.71 | 89.01 | **91.04** |
| *Kappa* | \ | \ | 81.31 | 79.29 | 83.93 | 82.79 | 81.24 | **85.29** |
| *FWIoU* | \ | \ | 80.66 | 80.84 | 82.65 | 81.71 | 80.44 | **83.77** |
| *Pred Time* | \ | \ | 262.47 s | **2.60 s** | 2.70 s | 513.43s | 3.21s | 4.02 s |

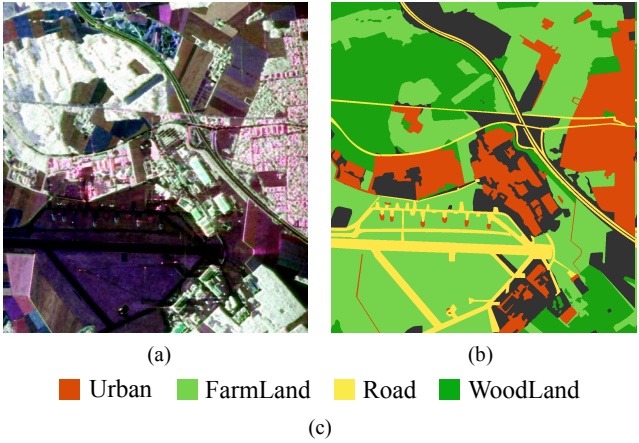

Fig. 8 *L*-band full PolSAR image data over the Oberpfaffenhofen region. (a) Pauli Pseudo RGB image. (b) Ground truth. (c) Label legends.

CRPM-Net achieves an increase of 1.3 percentage points on *OA* and a speed improvement of 24.5 times compared with Cs-CNN. Besides, it achieves more accurate classfication than RPM-Net but the speed is slightly lower. To sum up, CRPM-Net achieves the most acurate and fairly efficient classification on Oberpfaffenhofen, E-SAR dataset which includes many irregular terrains.

The confusion matrix of CRPM-Net is shown in Tab. IV. Same to the former experiment, result of roads is not very fine because of slim space shape and large amount of edges, and it is highly confused with farmland due to the same smooth surfaces and rough junction areas. Thus, the edge of the roads is easily affected by other classes shown in Fig. 9. At the same time, woodland is mainly confused with urban because of their irregular surface.

TABLE IV
CONFUSION MATRIX OF CRPM-NET MODEL ON OBERPFAFFENHOFEN IMAGE

| *Class* (%) | WoodLand | FarmLand | Urban | Roads |
|---|---|---|---|---|
| WoodLand | **91.76** | 2.04 | 5.82 | 0.38 |
| FarmLand | 0.52 | **94.60** | 2.25 | 2.63 |
| Urban | 3.32 | 5.44 | **89.43** | 1.81 |
| Roads | 3.02 | 14.98 | 5.34 | **76.66** |

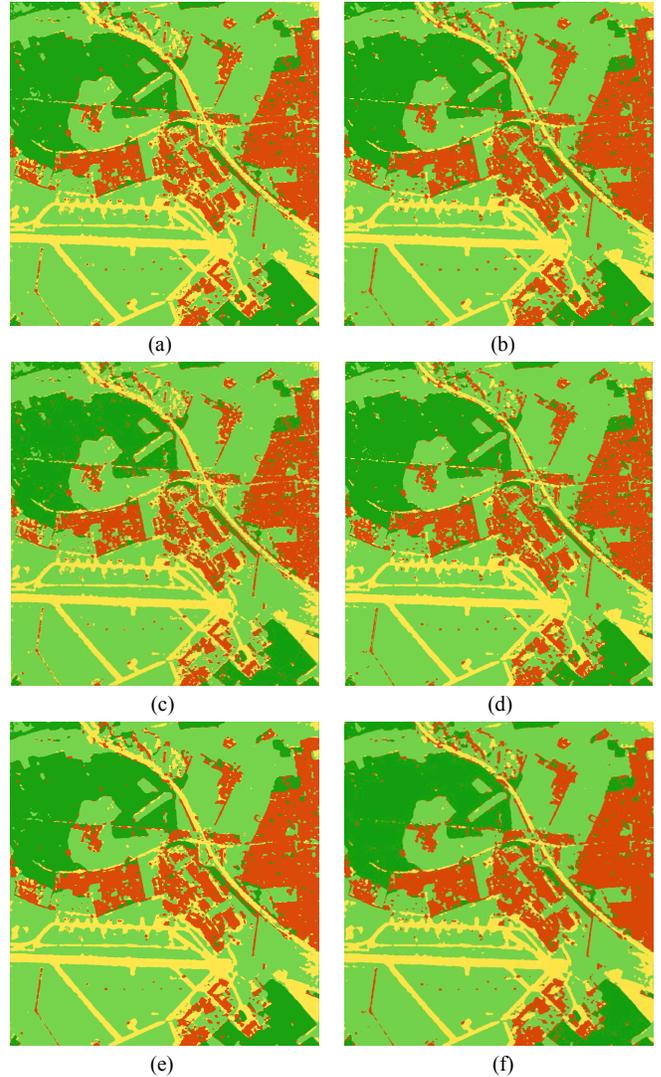

Fig. 9 Classification images of the Flevoland-Netherlands region. (a-f): CNN, Cs-CNN, Dilated CNN, C-Dilated CNN, RPM-Net, CRPM-Net.

The classification images of each model are shown in Fig. 9. There are many discrete pixels and holes in the result of Cs-CNN and C-Dilated CNN, at the same time, Fig. 9(f) shows that the number of holes and discrete pixels in CRPM-Net result image is much reduced, which is due to the contextuel semantic features and pixel-refining.

## C. Experiment on San Francisco Bay, AIRSAR, L-Band Dataset

The PolSAR data in the San Francisco Bay area is often used as experimental data for the terrain classification. It is the *L*-band single-look polarization data obtained by the AIRSAR platform. The image size is 900×1024, and its Pauli pseudo-color map is shown in Fig. 10(a), which contains a total of 5 terrain classes, including sea, urban, angled urban, vegetation and mountains. The ground truth is based on Google Earth in the San Francisco Bay area, shown in Fig. 10(b). The number of samples in each class is shown in Tab. V. Since the total number of samples of angled urban is small, the sampling rate is 5.6%, and sampling rates of other classes are all less than 2%.

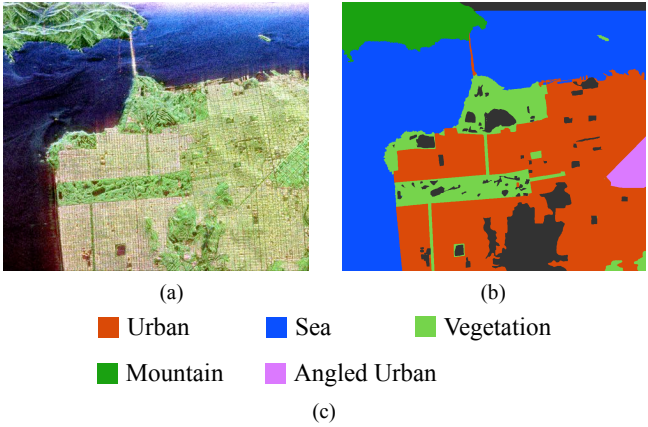

Fig. 10 *L*-band full PolSAR image data over the San Francisco Bay region. (a) Pauli Pseudo RGB image. (b) Ground truth. (c) Label legends.

Same with the experiment on Oberpfaffenhofen, the input channels of real and complex networks are 9 and 6, respectively. The evalutation results of CNN and Cs-CNN with 60 training epochs are shown in Tab. V, where the Cs-CNN increases the *OA* of CNN by 2.53%, the *Kappa* by 4.24%, and *FWIoU* by 3.43%. And after the parameters transferred to Dilated CNN and C-Dilated CNN, the *OA* decrease by 0.18% and 0.46%, respectively, whereas, the classification speed increase 75.3 times and 125 times, respectively. The last column in Tab. V shows that CRPM-Net is optimal for *OA*, *Kappa* and *FWIoU* evaluations, which are 96.21%, 94.01%, and 93.07%, respectively, moreover, the time cost is 100 times less than that of Cs-CNN.

From the confusion matrix of CRPM-Net shown in Tab. VI,

TABLE VI
CONFUSION MATRIX FOR OF CRPM-NET ON SAN FRANCISCO BAY IMAGE

| Class (%) | 1 | 2 | 3 | 4 | 5 |
|---|---|---|---|---|---|
| 1 Sea | **98.34** | 0.76 | 0.44 | 0.45 | 0.01 |
| 2 Vegetation | 0.62 | **83.34** | 2.62 | 9.97 | 3.45 |
| 3 Mountain | 1.0 | 1.94 | **96.84** | 0.17 | 0.05 |
| 4 Urban | 0.08 | 1.82 | 0.04 | **97.54** | 0.52 |
| 5 Angled Urban | 0.01 | 11.02 | 0.01 | 1.35 | **87.61** |

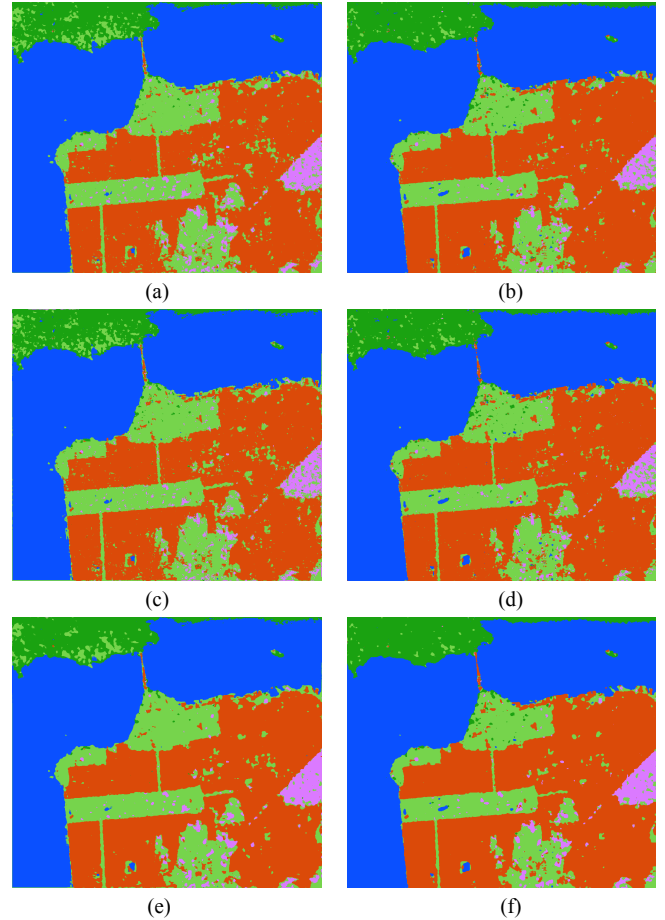

Fig. 11 Classification images of the San Francisco Bay region. (a-f): CNN, Cs-CNN, Dilated CNN, C-Dilated CNN, RPM-Net, CRPM-Net.

we can see that the main confusion occurs between angled urban and vegetation. Urban, sea, and mountain can be distinguished correctly for the huge texture feature differences.

TABLE V
CLASSIFICATION RESULTS OF SAN FRANCISCO BAY POLSAR IMAGE

| Class (%) | Train | Total | CNN | Dilated CNN | RPM-Net | Cs-CNN | C-Dilated CNN | CRPM-Net |
|---|---|---|---|---|---|---|---|---|
| Sea (0.3) | 1000 | 338685 | 97.53 | 97.08 | 97.23 | 98.39 | 98.39 | **98.34** |
| Vegetation (1.2) | 1000 | 84588 | 87.16 | 87.10 | **89.17** | 85.61 | 84.92 | 83.34 |
| Mountain (1.6) | 1000 | 63491 | 79.86 | 81.18 | 84.47 | 93.51 | 93.38 | **96.84** |
| Urban (0.4) | 1500 | 332040 | 92.85 | 92.96 | 95.10 | 96.52 | 96.14 | **97.53** |
| Angled Urban (5.6) | 800 | 14381 | 84.78 | 75.79 | **91.89** | 74.73 | 68.11 | 87.60 |
| **OA** | \ | \ | 93.07 | 92.89 | 94.51 | 95.60 | 95.14 | **96.21** |
| **Kappa** | \ | \ | 88.75 | 88.40 | 91.23 | 92.99 | 92.22 | **94.01** |
| **FWIoU** | \ | \ | 88.67 | 88.32 | 90.63 | 92.10 | 91.37 | **93.07** |
| **Pred Time** | \ | \ | 135.33 s | **2.03 s** | 2.07 s | 280.78 s | 2.24 s | 2.79 s |

The classification images are shown in Fig. 11. We can see that the classification result of Cs-CNN is more continuous in mountain, vegetation and urban areas than that of CNN. CRPM-Net can further improve the classification accuracy of mountains and urban.

D. *Experiment on Benchmark Flevoland AIRSAR, L-Band Dataset*

In order to verify the performance of CRPM-Net against some recent state-of-the-art classification algorithms, such as

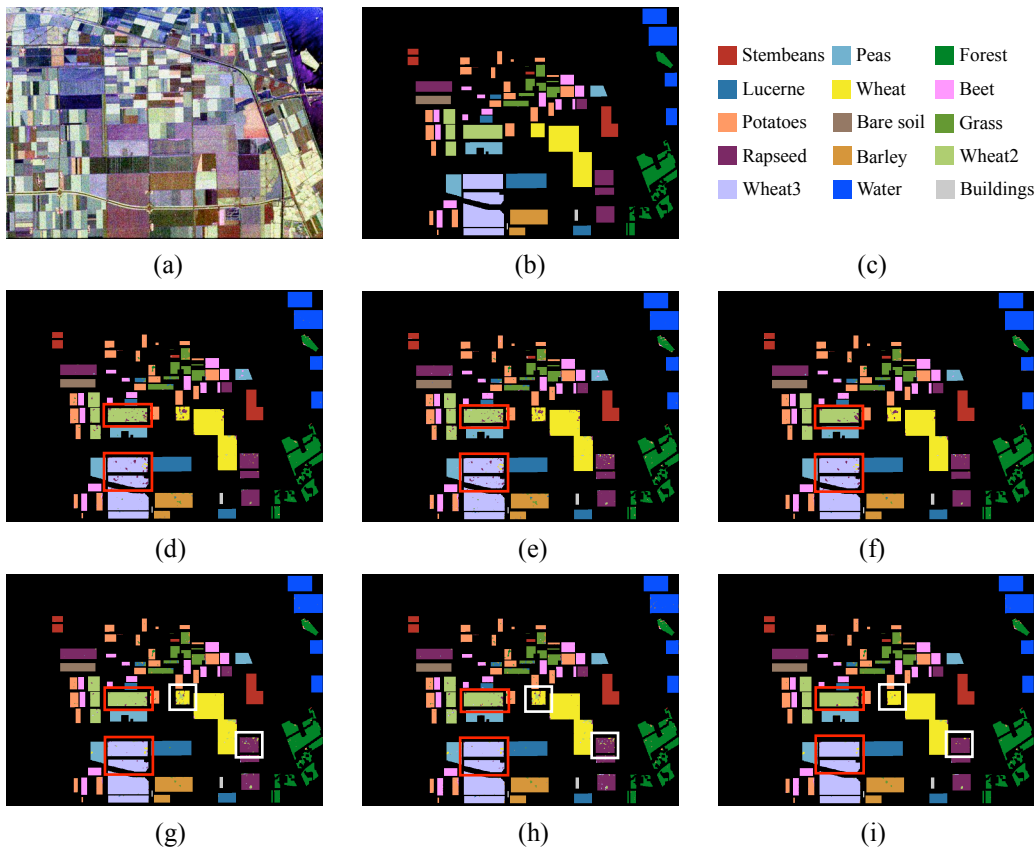

Fig. 12 *L*-band full PolSAR benchmark image data and its classification results images. (a) Pauli Pseudo RGB image. (b) Ground Truth. (c) Labels legends. (d-i): Classification results of CNN, Dilated CNN, RPM-Net, Cs-CNN, C-Dilated CNN and CRPM-Net.

TABLE VII
CLASSIFICATION RESULTS OF FLEVOLAND-NETHERLANDS POLSAR IMAGE

| Class (%) | Train | Total | CNN | Dilated CNN | RPM-Net | Cs-CNN | C-Dilated CNN | CRPM-Net | Liu *et al.* [33] PCN | Chen *et al.* [48] SAE |
|---|---|---|---|---|---|---|---|---|---|---|
| Stem beans (6.5) | 400 | 6103 | 99.66 | 99.09 | **99.71** | **99.71** | 99.54 | 99.70 | 96.59 | 98.30 |
| Peas (4.3) | 400 | 9111 | 98.62 | 98.32 | **99.32** | 99.28 | 98.17 | 98.43 | 95.40 | 96.47 |
| Forest (2.6) | 400 | 14944 | 98.43 | 97.32 | 98.33 | 98.48 | 97.80 | **98.65** | 95.11 | 98.09 |
| Lucerne (4.2) | 400 | 9477 | 98.71 | 98.62 | **98.90** | 97.30 | 98.06 | 98.46 | 93.67 | 97.78 |
| Wheat (2.3) | 400 | 17283 | **95.30** | 95.36 | 97.25 | 94.89 | 95.07 | 98.47 | 95.30 | **99.89** |
| Beet (3.9) | 400 | 10050 | 99.70 | 99.38 | **99.38** | 99.10 | 98.97 | 99.10 | 97.70 | 97.24 |
| Potatoes (2.6) | 400 | 15292 | 96.82 | 95.84 | **98.04** | 97.75 | 96.65 | 97.69 | 95.97 | 97.17 |
| Bare soil (6.4) | 200 | 3078 | 100.0 | 99.88 | **100.0** | **100.0** | **100.0** | **100.0** | 97.45 | 99.92 |
| Grass (6.4) | 400 | 6269 | 94.71 | 91.48 | 93.79 | 94.75 | 93.00 | **95.68** | 94.41 | 94.96 |
| Rapeseed (3.2) | 400 | 12690 | 98.47 | 97.39 | **98.78** | 96.85 | 96.27 | 98.52 | 94.05 | 96.03 |
| Barley (5.6) | 400 | 7156 | 96.32 | 95.64 | 97.82 | 98.49 | 98.39 | 98.50 | 94.60 | **99.17** |
| Wheat 2 (3.8) | 400 | 10591 | 93.64 | 93.00 | 95.85 | 97.71 | 96.14 | **97.93** | 95.13 | 95.44 |
| Wheat 3 (1.9) | 400 | 21300 | 95.53 | 95.81 | 97.17 | 98.14 | 98.43 | **99.12** | 95.51 | 98.50 |
| Water (2.9) | 400 | 13476 | 99.70 | 99.73 | 99.83 | 99.73 | 99.21 | 99.83 | 99.65 | **99.89** |
| Buildings (8.4) | 40 | 476 | **100.0** | 100.0 | 100.0 | 100.0 | 100.0 | 100.0 | 95.61 | 92.11 |
| **OA** | \ | \ | 97.27 | 96.75 | 98.07 | 97.86 | 97.42 | **98.60** | 97.07 | 97.62 |
| **Kappa** | \ | \ | 96.70 | 96.43 | 97.88 | 97.65 | 97.16 | **98.29** | 94.15 | 97.41 |
| **FWIoU** | \ | \ | 94.91 | 93.90 | 96.35 | 95.89 | 95.03 | **96.98** | \ | \ |
| **Pred Time** | \ | \ | 114.45 s | 1.73 s | 2.10 s | 237.86 s | **1.58s** | 2.45 s | 35 s | 80.70 s |

[33] and [48], we use a wildly studied *L*-band full PolSAR image over Flevoland as the benchmark data for cross evaluation. The image size is 750×1024 pixels, and a pseudo RGB image formed by the Pauli decomposition is shown in Fig. 12(a). The ground truth [49] and label legends are shown in Fig. 12(b) and (c). 15 classes of terrains are labeled in this image, they are wheat, beet, potatoes, bare soil, grass, rapeseed, barley, water, and buildings. The number of training pixels and total pixels for each class is shown in the second and third cols in Tab. VII, respectively. Among the 15 classes, buildings have extremely few pixels in ground truth, so we just selected 40 pixels for building class randomly. Sampling rates of each class are shown in the first col in Tab. VII.

The complex domain networks can recognize rapeseed, wheat2 and wheat3 much better than real domain networks [as highlighted by the two red rectangles in Fig. 12(d), (g)]. CRPM-Net improves the performance on rapeseed greatly compared to Cs-CNN and RPM-Net, which is clearly illustrated by the white rectangles in Fig. 12(g), (h) and (i). Moreover, according to the *OA*, *Kappa* coefficients, *FWIoU*, and time cost shown in Tab. VII. CRPM-Net achieves the highest score among algorithms evaluated at a fairly fast speed.

Moreover, two recently published state-of-the-art studies are cross compared in 9-th, 10-th col in Tab. VII. Liu *et al*. [33] proposed the PCN and obtained 97.07% accuracy with more sampled pixels. Chen *et al*. [48] used SAE-SDPL algorithm to achieve 97.62% accuracy with 80.7 seconds. Obviously, CRPM-Net gets the best performance on *OA*, *Kappa* coefficients and time cost which validates its effectiveness.

*E. Analysis of results*

*1) Accuracy*: The *OA*, *Kappa*, *FWIoU* and Time cost results of the four experiments are shown in Fig. 13. As it can be seen from (a), (c) and (d), Complex cross-convolution can effectively improve the classification accuracy of real domain networks, which is mainly due to the full use of polarimatric target decomposition features and phase information. Whether in real domain or complex domain, the classification accuracy of Dilated CNN drop slightly with transferring parameters directly from CNN, but the classification speed increased dramatically as shown in Fig. 13(b) for less repeated calculation. Most importantly, the CRPM-Net proposed in this paper achieves the best results in *OA*, *Kappa*, *FWIoU* and Time cost.

*2) Efficiency*: Fig. 13(b) shows that the classification time of each network on the whole PolSAR image. As for CNN and Cs-CNN, the whole image classification is implemented by row-by-row classification [batchsize=row length], so the running time is 80 times longer than those of RPM-Net and CRPM-Net. In addition, CNN is 2 times faster than Cs-CNN for the double number of multiplications in complex cross-convolution in Cs-CNN. As for CRPM-Net and C-Dilated CNN, CRPM-Net is more accurate but slightly slower for the parallel networks.

In conclusion, CRPM-Net achieves the most accurate and fairly fast classification on various PolSAR images from different platforms owing to the consideration of contextual semantic features and precise localization.

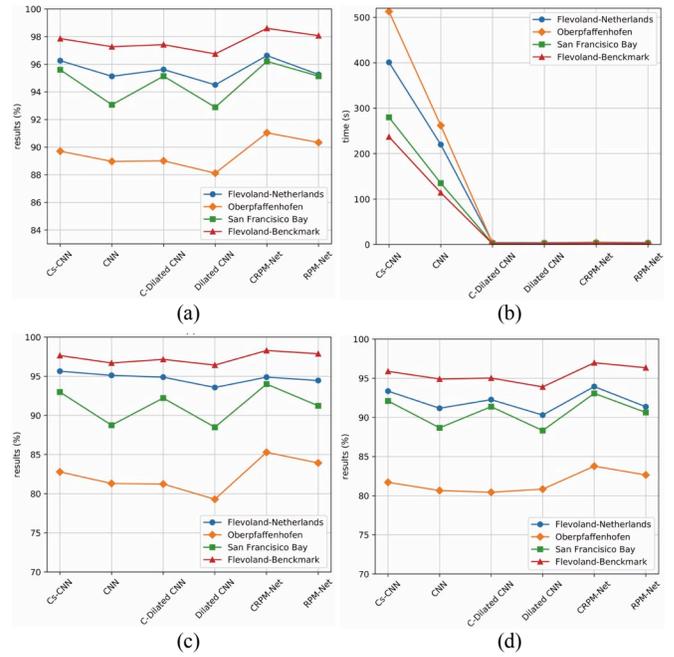

Fig. 13 Classification results of four experiments. (a-d): Overall Accuracy, Time cost, *Kappa*, *FWIoU*.

## V. CONCLUSION

In this paper, we propose an efficient and accurate PolSAR image classification algorithm based on dilated convolution and pixel-refining parallel mapping network in the complex domain, i.e., CRPM-Net. In proposed algorithm, a complex cross-convolution network (Cs-CNN) is trained with a small number of labeled pixels at first, aiming to extract high dimension features from polarimetric target decomposition features and phase information. Then, we transfer the parameters of Cs-CNN to a dilated convolution neural network in the complex domain (C-Dilated CNN) for direct pixel mapping, which can achieve huge improvement in efficiency bearing only a slight decrease in accuracy for all experiments. With a parallel concatenation with C-Dilated CNN and an encoder-decoder network in the complex domain, we obtain the proposed CRPM-Net, which can not only enable good localization but also capture contextual semantic features. Finally, after trained with the fusion of dense score map from C-Dilated CNN and high-weighted training pixels, CRPM-Net achieves a more than 80 times speed improvement compared to Cs-CNN with better results on *OA*, *Kappa*, and *FWIoU*. In addition, CRPM-Net further reduces the number of holes and discrete areas so that the region of each class is more coherence.

Compared with the networks in real domain, the proposed cross-convolution can achieve better classification results for full use of polarimetric target decomposition features and phase information.

In conclusion, CRPM-Net achieves the best classification results in all experiments and substantially outperforms some latest state-of-the-art approaches in both efficiency and accuracy with a small number of labeled pixels.

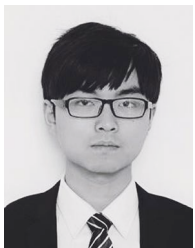
**DongLing Xiao** was born in China in 1995. He received the B.E. degree in electronic engineering from University of Electronic Science and Technology of China, Chengdu, China, in 2016. He is currently pursuing the M.S. degree with the Department of Airborne Microwave Remote Sensing System, Institute of Electrics, Chinese Academy of Sciences, Beijing, China.

His research interests include pattern recognition, machine learning, remote sensing image processing, and deep learning.

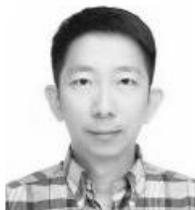
**Chang Liu** received the B.E. degree in electronic engineering from Tsinghua University, Beijing, China, in 2000, and the M.S., E.E., and Ph.D. degrees in electronic engineering from Institute of Electrics, Chinese Academy of Sciences, Beijing, China, in 2006.

His research interests include multi-function Synthetic Aperture Radar system, high performance embedded radar signal processing, SAR image application. He was awarded the second-class Prize of The State Scientific and Technological Progress Award, in 2010 and 2016 respectively

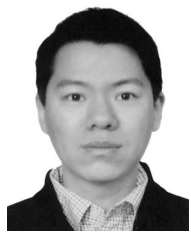
**Qi Wang** was born in China in 1976, He received the B.E degree from the Department of Automation in Tsinghua University in 1999, and PhD degrees in University of Chinese Academy of Sciences, Beijing, China, in 2005. Currently, he is working in the Department of Airborne Microwave Remote Sensing System, Institute of Electrics, Chinese Academy of Sciences, Beijing, China.

His research interests include the radar imaging algorithm, target detecting, tracking and classification.

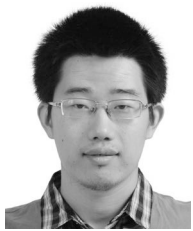
**Chao Wang** was born in China in 1983. He received the B.E. degree in electronic engineering from Tsinghua University, Beijing, China, in 2006, and the M.S. degree in computer application from University of Chinese Academy of Sciences, Beijing, China, in 2009. He is currently pursuing the Ph.D. degree with the Key Laboratory of Information Science of Electromagnetic Waves, Institute of Electrics, Chinese Academy of Sciences, Beijing, China.

His research interests include signal processing, remote sensing image processing, and deep learning.

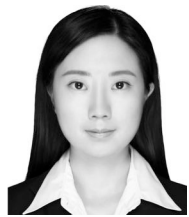
**Xin Zhang** was born in China in 1992. She received the M.S. degree in cartography and geographic information systems from China Agricultural University, China, in 2018. Currently, she is working in Department of Airborne Microwave Remote Sensing System, Institute of Electrics, Chinese Academy of Sciences, Beijing, China.

Her research interests include the remote sensing image processing and deep learning.